# Vision Based Navigation for a Mobile Robot with Different Field of Views

Rizwan A. Khan, M. Aasim Qureshi, Saqib Saeed
rizkhan@gmail.com, maasimq@hotmail.com, saqib.saeed@gmail.com
Bahria University, Islamabad.

## ABSTRACT

The basic idea behind Evolutionary Robotics is to evolve a set of neural controllers for a particular task at hand. It involves use of various input parameters such as Infra-red sensors, light sensors and vision based methods. This paper aims to explore the evolution of vision based navigation in a mobile robot. It discusses in detail the effect of different field of views for a mobile robot. The individuals have been evolved using different FOV values and the results have been recorded and analyzed. The optimum values for FOV have been proposed after evaluating more than 100 different values. It has been observed that the optimum FOV value requires lesser number of generations for evolution and the mobile robot trained with that particular value is able to navigate well in the environment.

Keywords: Artificial Neural Networks, Genetic Algorithms, Autonomous Robots, Vision

## 1 Introduction

Autonomous Robots has been an idea taken from Braitenberg's *Vehicles* where it is suggested that Evolutionary techniques be used to develop mobile robots. Rather than attempting to hand-design a system to perform a particular task or range of tasks well, the evolutionary approach allows a gradual emergence of the sought after behavior.

Though more closely related to Behavioral Robotics than to Classic Robotics, Evolutionary Robotics is a third technique of robotics; it uses neither functional nor behavioral decomposition. Given a suitable control structure (brain), it is utterly up to evolution to figure out the best way to solve the task. It is possible for the designer to split a large task into subtasks and evolve them.

In this paper we describe the evolution of a discrete time recurrent neural network which aims to move in an environment avoiding obstacles using its vision capabilities. Some work has been done in this regard at EPFL Switzerland by Dr. Dario Floreano [1], [2], [4], [5] and his team and also at Sussex University [6].

The aim of this work is to study and analyze the effects of changes that occur to the evolutionary mechanism if we change the environment characteristics. Various changes are also made in the genetic evolution parameters like FOV and the results of different experiments are compared. All of the experiments have been done using the Teem [3] Simulator which is freely available for research in Evolutionary Robotics.

## 2 The Experiments

The aim of the experiments is to find the effect of Field of View and No. of Generations on the fitness values of the neural controllers being evolved for a mobile robot. A very simple environment has been used with only the walls acting as obstacles for the robot. To make a complete analysis of the whole task, almost 200 MB of data was gathered over a period of more than 100 consecutive hours of different simulations.

All the experiments use Khepera [7] as the mobile robot which is equipped with a linear vision camera. Its configuration can be varied as desired in the simulation. The input that the camera received in this case is processed rather than raw pixel data. The environment used is a simple one with walls acting as the main obstacles in the rectangular area.

## 3 Results

To get a detailed picture, we simulated our robot for Field Of View (FOV) ranging from 0 to 180 degrees and evolved the controller for 100 generations having 60 individuals in each generation. To nullify the random effect and get stable data for analysis the evolution runs were repeated for 5 times for each FOV and then the final results were based on the mean taken from those multiple runs. The fitness graph for the average individual in each generation and for each FOV value is shown in fig.1 while fig.2 shows the same fitness values graph for best individuals.

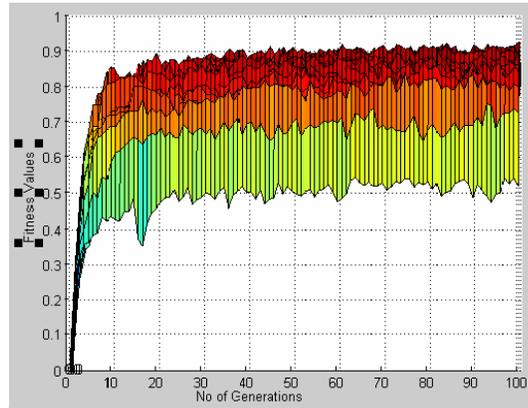

**Figure 2: Fitness vs. No Of Generations**

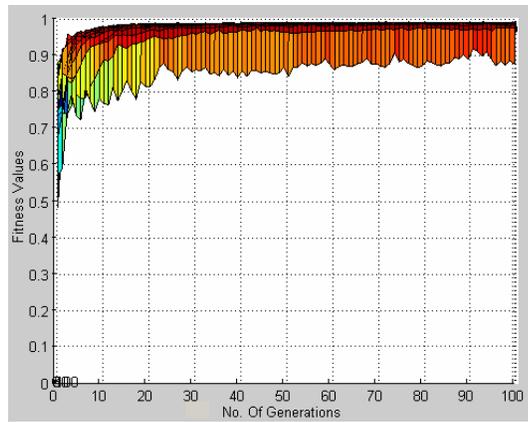

**Figure 1: Fitness vs. No of Generations**

As evident form the fig.1 & fig.2, it is noted that almost in each case the learning process stabilizes within 30 generations. So in order to get more useful data, and to have a closer look at it, we reduced the number of generations to 30. It still incorporates a huge number of around 9000 (1800 x 5) individuals being tested. We analyze the data on the basis of these three parameters for an average (average individual of a generation is constructed from the mean of 60 individuals of that generation) and best individual:

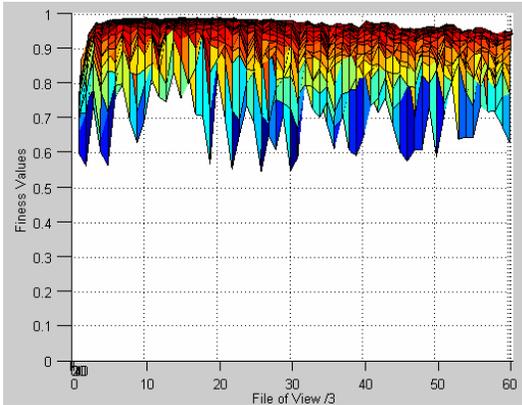

**Figure 3: Fitness vs. Field of View of the Best Individual**

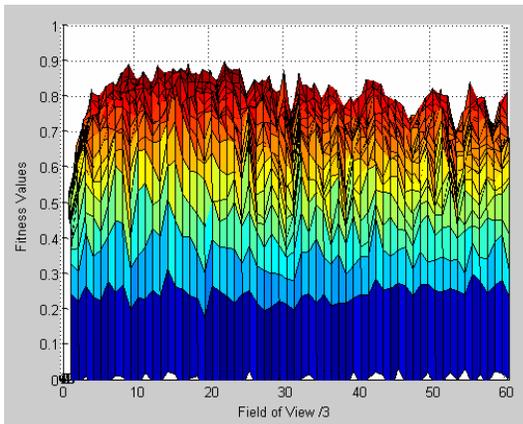

**Figure 4: Fitness vs. Field of View of an Average Individual**

1. Relation between FOV and Fitness Values.
2. Relation between No. of Generations and Fitness Values for a particular FOV.
3. Relation between FOV and No. of Generations to achieve a specific fitness value.

### 3.1 FOV and Fitness Values

It was noted that an individual having a fitness value of 0.85 or above is considered to be a good individual. So, on the basis of this observation, it was found that the fitness value is above 0.85 mark when and average individual is with a FOV in the range 15 degree to 75 degree (see fig.3). In case of the best individual, the fitness value rises to 0.95 for the same range of FOVs (see fig.4). The performance of individuals having an FOV value less than 15 is not satisfactory and an average individual has a low fitness value of 0.5, and it decreases sharply as the FOV value decreases. For FOV values greater than 75, the robot behaves satisfactorily in the case of best individual but the behavior is not promising for an average individual.

### 3.2 No. of Generations and Fitness Values for a particular FOV

In this comparison we studied the impact of number of generations on the training of an individual. As it is obvious from fig.5, for most of the individuals fitness value reaches to its maximum by 15 – 20 generations. Fitness improves as number of generations increases but after reaching a specific point it stabilizes that shows the capability of that individual and after that the improvement tends to be very slow. In fig.5 we can see that most of the average individuals reach a max point and then stabilize. The same thing can be observed for the set of best individuals from each generation. The trend of the graph in both of the cases shows the same inclination and effect but with a difference in fitness. In fig.5 most of the average individuals stabilize at fitness value of .85 after 12 to 15 generations and as shown in fig.6 the best individuals of each generation stabilize little early at fitness value around .98 after 8 to 10 generations.

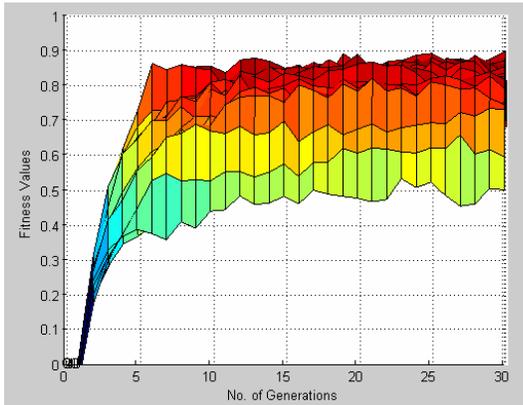

**Figure 5: Fitness vs. Generations of an Average Individual**

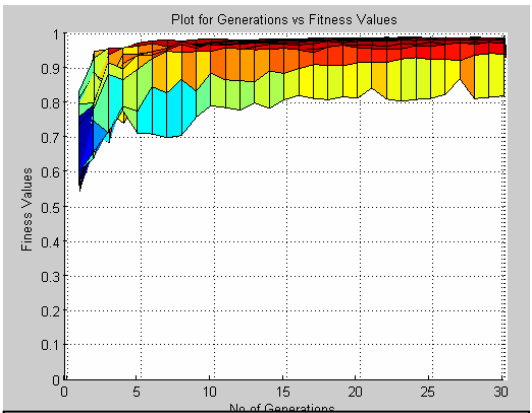

**Figure 6: Fitness vs. Generations of the Best Individual**

### 3.3 No. of Generations and Field of View

In fig.7 and fig.8 the colors show the intensity at different points in 2-D graph. Ranges from dark blue to maroon – number of generation required to bring an individual to its best. The maroon color shows the best performance (that is good fitness values). Here in fig.7 it can be seen that there are too few maroon spots and most of them are clustered together for the FOV in the range 10 – 90 where the learning process is at its best. This shows that the learning process during training is the fastest when FOV is between 10 and 90. Moreover, the marked area (encircled with black line) in fig.7 shows that the highest fitness value is for FOV close to 45 degrees. The same trend can be seen in the case of best individuals (see fig.8), where we see the same range of FOV values but good fitness values are relatively more scattered in this case. Here again it is seen that the best fitness values are concentrated very close to the FOV value of 45 degrees as shown by the marked area (encircled with black line) in fig.8. Here in this scenario we can see few maroon spots scattered here and there but they are too few to be brought into study.

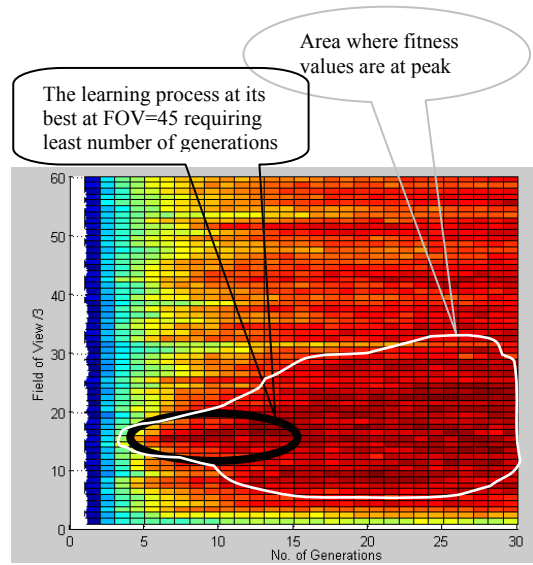

**Figure 7: Field of View vs. Generations for all fitness values of an Average Individual**

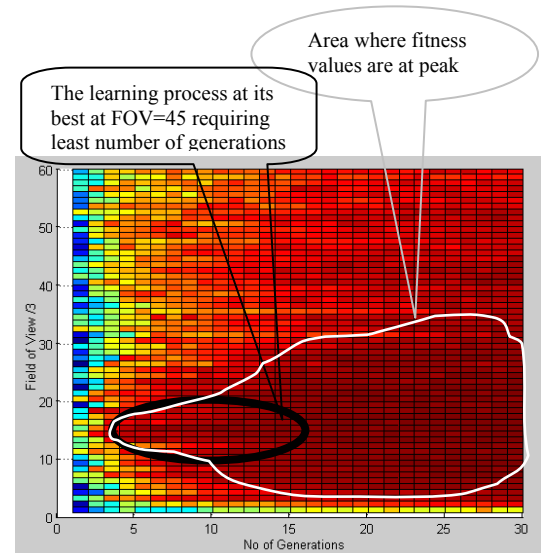

**Figure 8: Field of View vs. Generations for all fitness values of the Best Individual**

## 4 Conclusions

The above study provided us with following results:
- Field of view has a deep effect on both number of generations and fitness value and the best Field of View is 45 degrees.
- More than 95 % of individuals stabilize to their best fitness value by $30^{th}$ generation.